\documentclass[10pt,twocolumn,letterpaper]{article}

\usepackage{cvpr}              
\usepackage{graphics}
\usepackage{graphicx}
\usepackage{amsmath}
\usepackage{amssymb}
\usepackage{booktabs}
\usepackage{arydshln}
\usepackage{times}
\usepackage{epsfig}
\usepackage{etoolbox}
\usepackage[export]{adjustbox}
\usepackage{framed}
\usepackage{color, colortbl}
\usepackage{setspace}
\usepackage{multirow}
\usepackage{placeins}
\usepackage{cuted}
\usepackage{capt-of}
\usepackage{pifont}
\usepackage{enumitem}
\usepackage{bbm}
\usepackage{dsfont}
\usepackage{algorithm}
\usepackage{algpseudocode}
\usepackage[table]{xcolor}

\definecolor{cvprblue}{rgb}{0.21,0.49,0.74}
\usepackage[pagebackref,breaklinks,colorlinks,citecolor=cvprblue]{hyperref}

\def\paperID{26} 
\def\confName{CVPR}
\def\confYear{2024}

\title{Performance Evaluation of Segment Anything Model with Variational Prompting for Application to Non-Visible Spectrum Imagery
\vspace{-.0cm}}
\author{Yona Falinie A. Gaus\textsuperscript{1}, Neelanjan Bhowmik\textsuperscript{1}, Brian K. S. Isaac-Medina\textsuperscript{1}, Toby P. Breckon\textsuperscript{1, 2}\\
Department of \{\textsuperscript{1}Computer Science, \textsuperscript{2}Engineering\}, Durham University, UK
}

\begin{document}
\maketitle

\begin{strip}
\centering
\includegraphics[width=\textwidth]{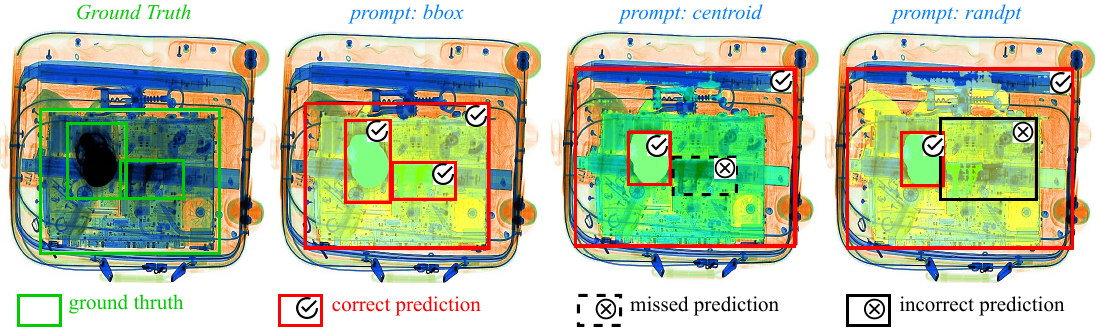}
\vspace{-0.4cm}
\captionof{figure}{We propose evaluating three prompting strategies (bounding box - \textit{bbox}, \textit{centroid}, random point - \textit{randpt}) to assess the effectiveness of the Segment Anything Model applied to X-ray and infrared imagery for identifying objects of interest. The bbox prompt yields superior segmentation results, while the other two prompting strategies demonstrate notably higher incorrect/missed predictions. 
\label{fig:teaser}}
\end{strip}

\begin{abstract} \label{sec:abs}
\vspace{-0.4cm}
\noindent
The Segment Anything Model (SAM) is a deep neural network foundational model designed to perform instance segmentation which has gained significant popularity given its zero-shot segmentation ability.SAM operates by generating masks based on various input prompts such as text, bounding boxes, points, or masks, introducing a novel methodology to overcome the constraints posed by dataset-specific scarcity. While SAM is trained on an extensive dataset, comprising more than $11M$ images, it mostly consists of natural photographic (visible band) images with only very limited images from other modalities. Whilst the rapid progress in visual infrared surveillance and X-ray security screening imaging technologies, driven forward by advances in deep learning, has significantly enhanced the ability to detect, classify and segment objects with high accuracy, it is not evident if the SAM zero-shot capabilities can be transferred to such modalities beyond the visible spectrum. For this reason, this work comprehensively assesses SAM capabilities in segmenting objects of interest in the X-ray and infrared imaging modalities. Our approach reuses and preserves the pre-trained SAM with three different prompts, namely bounding box, centroid and random points. We present several quantitative and qualitative results to showcase the performance of SAM on selected datasets. Our results show that SAM can segment objects in the X-ray modality when given a box prompt, but its performance varies for point prompts. Specifically, SAM performs poorly in segmenting slender objects and organic materials, such as plastic bottles. Additionally, we find that infrared objects are also challenging to segment with point prompts given the low-contrast nature of this modality. Overall, this study shows that while SAM demonstrates outstanding zero-shot capabilities with box prompts, its performance ranges from moderate to poor for point prompts, indicating that special consideration on the cross-modal generalisation of SAM is needed when considering use on X-ray and infrared imagery. 

\end{abstract}    
\section{Introduction}
\label{sec:intro}

In the domain of security, the strategic deployment of advanced imaging technologies holds pivotal significance, contributing significantly to safeguarding national borders, airports, public facilities, transportation systems, and national infrastructure.
Infrared-band camera imagery is well established within visual surveillance, offering extensive applications in target detection, visual tracking, behaviour analytics, home monitoring, and automotive environment perception \cite{smeulders2013visual,kundegorski2014photogrammetric,kundegorski2016real,kundegorski2015posture, li2010robust,teutsch2014low, brehar2014pedestrian}. 
In the domain of X-ray imaging, X-ray security screening stands as a widely utilised method in aviation and broader transportation sectors, detecting prohibited items by scrutinising X-ray images of baggage, freight, and postal items \cite{webb2021operationalizing, gaus2019evaluation, miao19:sixray}.


The recent rise of Convolutional Neural Networks (CNN) \cite{krizhevsky2012imagenet} has revolutionised visual tasks significantly advancing the state-of-the-art in image understanding technologies. 
Within object detection, most efforts have focused on detecting objects-of-interest in standard colour imagery  by  using multi-stage \cite{ren17:fasterrcnn,MaskRCNNHe2017, cai19:cascade}, single-stage \cite{redmon2018yolov3, lin18:retinanet, zhu19:fsaf} and transformer-based \cite{carion20:detr, zhu2021:deformable} detectors. These aforementioned object detection based CNN methods rely heavily on architectures that have been trained on large-scale colour imagery datasets such as ImageNet \cite{deng2009imagenet}.  Introducing CNN to object detection within infrared and X-ray imagery is significantly hindered by the absence of such annotated datasets of the same scale and variety \cite{kundegorski16xray, turcsany13xray, mery13:xray}. 

To address the challenges in the field of infrared imagery analysis, methods such as transfer learning \cite{gaus2020visible}, generation of pseudo-RGB equivalents \cite{devaguptapu2019borrow}, and domain adaptation \cite{munir2021sstn} have been utilised to enhance existing CNN models and establish public benchmarks for research. Gaus et al. \cite{gaus2020visible} concentrate on detecting objects in infrared imagery by leveraging a transfer learning approach, where the knowledge obtained from the visible spectrum is transferred to the infrared domain. Devaguptapu et al. \cite{devaguptapu2019borrow} employ image-to-image translation techniques to create pseudo-RGB versions of infrared images. These pseudo-RGB are then processed using CNN models for object detection in infrared imagery. Munir et al. \cite{munir2021sstn} introduce a method of self-supervised domain adaptation through an encoder-decoder transformer network to develop a robust infrared image object detector in autonomous driving.

This methodology parallels efforts in the broader domain of X-ray security imaging, where several benchmarks for security inspection have been developed \cite{mery2015gdxray,miao2019sixray,zhang2022pidray,wei2020:opixray}. Public datasets such as GDXray \cite{mery2015gdxray}, SIXray \cite{miao2019sixray}, PIDray \cite{zhang2022pidray} or OPIXray \cite{wei2020:opixray} have been released, where the main goal is to advance the developments of prohibited item detection in X-ray images using CNN based methods \cite{akcay2018using,chang2022detecting,gu2020automatic, liu2018detection, subramani2020evaluating,webb2021operationalizing}.

The performance of these CNN-based models heavily depends on the availability of suitable infrared and X-ray imagery datasets with sufficient object annotations, diversity and scale, which has often been lacking compared to visible imagery resources. A common solution to address this issue involves pre-training, which effectively utilizes a limited volume of target dataset (X-ray and infrared imagery). However
pre-training on datasets beyond visible imagery could lead to dataset bias, potentially misaligning with the idiosyncratic characteristics of the target dataset, which significantly diverges from visible datasets \cite{isaac2023seeing}. For example, X-ray datasets consist of semi-transparent transmission imagery, where objects appear translucent and blend visually from front to back, unlike visible images where foreground objects occlude those in the background. Conversely, infrared images are not influenced by variations in visible spectrum illumination and shadows, illustrating the unique challenges and differences each imaging modality presents when compared to standard visual datasets. 
To address these challenges, the development of comprehensive datasets enriched with detailed annotations in non-visible spectrum imagery becomes essential. This approach serves not only to enhance model training and performance but also to ensure broader applicability and effectiveness across varying imaging modalities. In this context, developing foundation models \cite{bommasani2021opportunities}  and zero-shot learning \cite{xian2018zero} techniques can significantly alleviate the common challenges for datasets of different modalities. Foundation models are neural networks that undergo training on an extensive body of data, utilising innovative learning and prompting strategies that generally bypass the need for conventional supervised training labels. This approach enhances their capability to apply zero-shot learning to entirely new datasets across diverse settings. 

Whilst foundation models have revolutionised the field of natural language processing \cite{devlin2018bert,mann2020language,radford2021learning}, the Segment Anything Model (SAM) \cite{kirillov2023segment} has demonstrated promising zero-shot segmentation capabilities across multiple datasets of natural images. Therefore, to address the issue of demanding requirement for extensive annotated datasets in non-visible spectrum, this work examines the application of SAM and its effects on identifying objects of interest under X-ray (PIDray \cite{ma2022:pidray}, CLCXray \cite{zhao22:clcxray}, DBF6 \cite{akcay2018ganomaly}) and infrared imagery datasets (FLIR \cite{flir2019flir}). Utilising the variational prompting capabilities of SAM (bounding box, centroid, and random points), we conduct a thorough quantitative and qualitative analysis of the segmentation results produced by SAM (Fig. \ref{fig:teaser}). We aim to pave the way for utilising SAM to enhance the segmentation of object-of-interest beyond visible spectrum imagery through this evaluative research.

\begin{figure*}[ht!]
\centering
\subfloat{\includegraphics[width=\linewidth]{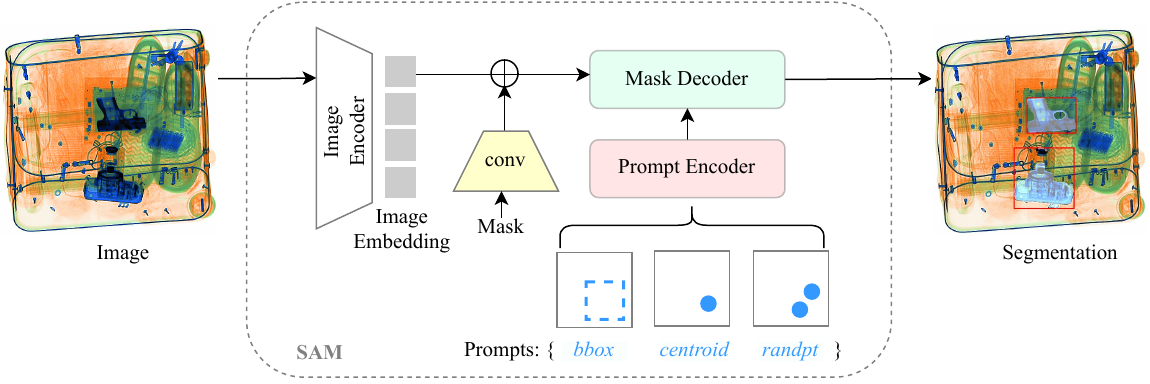}}
\vspace{-0.3cm}
\caption{Given an input image, Segment Anything Model (SAM) initiates the process by generating image embeddings via an image encoder. These embeddings are subsequently interactively queried by variational prompts ({bounding box, centroid, and random points}) in order to generate precise segmentation masks for the objects of interest.}
\label{fig:full_arch_}
\end{figure*}

\section{Literature Review}\label{sec:lr}

\noindent
Research works on non-visible spectrum imagery have witnessed increased attention in the literature. In this context, infrared imaging is progressively gaining traction across various fields, driven by the decreasing size and cost of its sensors. This trend has positioned it as a preferred choice for applications in visual surveillance and autonomous driving \cite{gaus2020visible, devaguptapu2019borrow, peng2016nirfacenet, lee2016recognizing, rodger2016classifying, bhowmik2022lost}. Moreover, there has been a steady rise in research focused on object-based prohibited items \cite{webb2021operationalizing, bhowmik19synthetic, bhowmik21energy} and anomaly detection \cite{gaus2019evaluation, Bhowmik2019, akcay2018ganomaly} within X-ray baggage security imagery. For a comprehensive review of X-ray security screening, readers can refer to the works of \cite{akcay22survey, mehdi23:xraysurvey, div22:xraysur}.

Key public datasets support the utilisation of CNN-based object detection beyond the visible spectrum. For Infrared imagery, pivotal resources include the FLIR ADAS dataset \cite{flir2019flir} and the Multispectral KAIST dataset \cite{hwang2015multispectral}, while for X-ray imaging, popular datasets such as GDXray \cite{mery2015gdxray}, SIXray \cite{miao2019sixray}, PIDray \cite{zhang2022pidray}, and OPIXray \cite{wei2020:opixray} play a crucial role.
Both datasets are designed to emphasize the advancement of CNN-driven object 
detection systems by offering detailed annotations beyond the visible spectrum in both Infrared and visible imagery.

The effectiveness of object detection methods depends significantly on the presence of annotated and labelled Infrared and X-ray images. Such a universal model can be achieved via foundational models such as SAM, showcasing its remarkable performance across various medical segmentation tasks \cite{ma2024segment,mazurowski2023segment, cheng2023sam, huang2024segment}.
The investigations into SAM performance across a diverse array of medical imagery demonstrated that while SAM achieves commendable segmentation results on targets with clear boundaries, it struggles significantly with typical medical subjects that have weak boundaries or exhibit low contrast \cite{mazurowski2023segment,huang2024segment}. 
While prior methods utilised the standard SAM directly for segmentation tasks, MedSAM \cite{ma2024segment} has adopted a distinct strategy by fine-tuning SAM on a large dataset containing over one million medical image-mask pairs.
The results demonstrate that MedSAM significantly enhances the segmentation performance and outperforms specialist models that were trained from the same modality \cite{isensee2021nnu}.

Drawing on the success of SAM in medical imagery, this study seeks to provide a thorough examination into the efficacy of SAM in segmenting imagery of non-traditional modalities, specifically Infrared and X-ray images, without the need for re-training or fine-tuning. Our study explores the zero-shot capabilities of SAM across three public X-ray datasets \cite{zhao22:clcxray,zhang2022pidray,akcay2018using} targeted on prohibited items such as firearms, knives, hammers, etc., alongside one public infrared dataset \cite{flir2019flir} where the detected objects include pedestrians, cars, bicycles, and similar entities. We aspire that this preliminary investigation offers insights into the performance of SAM beyond the visible spectrum, potentially looking into its applicability for generating high-quality annotations in infrared and X-ray imagery. Our goal is to evaluate whether SAM could facilitate the curation and offer detailed annotation of new datasets beyond the visible spectrum, fostering further advancements in the field.

\section{Methodology}
\label{sec:method}
 We propose SAM to produce high-quality zero-shot segmentation masks for datasets beyond the visible spectrum.  In Section \ref{sec:sam_arch}, we first briefly review the architecture of SAM, followed by Section \ref{sam_app} which addresses our primary application of prompting capabilities of SAM to generate segmentation masks for infrared and X-ray modality images.

\subsection{SAM architecture}
\label{sec:sam_arch}
The Segment Anything Model (SAM) is a foundation model that has achieved promising zero-shot segmentation performance, trained on a large visible imagery dataset. It is done by isolating specific objects within an image based on user-defined prompts. These prompts can vary from a single point, a full mask, a bounding box or text.
SAM mainly consists of three modules, as depicted in Fig. \ref{fig:full_arch_}. The first module, the image encoder, is composed of a Vision Transformer (ViT) \cite{dosovitskiy2020image} backbone for image feature extraction, resulting in image embedding in a spatial size of $64\times64$. The second module, the prompt encoder, encodes the interactive positional information derived from input points, boxes, or masks, to provide for the mask decoder. The third module, the mask decoder, consists of a two-layer transformer-based decoder which takes both the extracted image embedding with the concatenated output and prompt tokens for final mask prediction. The core principle of SAM lies in its ability to show strong zero-shot generalisation to new data without the necessity for additional training, since it was trained progressively on the large-scale Segment Anything $1$ Billion (SA-1B) dataset, which contains over $1$ billion automatically generated masks ($400\times$ more masks than any existing segmentation datasets \cite{gupta2019lvis, kuznetsova2020open}) and $11$ million images. 

\subsection{SAM application}
\label{sam_app}
A key feature of SAM in the second module, as explained in Section \ref{sec:sam_arch}, is the selection of the appropriate segmentation prompts (Fig. \ref{fig:full_arch_}). While automatic mask generators without manual prompts can be derived, this work focuses on isolating only particular objects of interest, rather than segmenting every object present.  Therefore, an auto-prompt approach does not align with this task.

We propose that SAM can be employed through two different prompting conditions for segmenting objects of interest beyond the visible spectrum. First, we use points as prompts. In this setup, a series of specific points within the object of interest in the image is provided to guide the processing of SAM. We provide two types of point prompts. The first type is \textit{centroid}, where we defined the point as the centre of the mask given at each object. The second type is \textit{randpt}, where we defined the prompt as two random points inside the mask given at each object. In addition, bounding box prompts are tested as input prompts for each object of interest, akin to security inspections. This is conducted by using the ground truth bounding box given for each image. The prompting techniques used in this work are:  
\begin{itemize}[noitemsep,topsep=0pt,leftmargin=*]
    \item \textbf{SAM-bbox}: where we employed ground truth bounding box as prompt. 
    \item \textbf{SAM-centroid}: where we defined SAM-centroid as the mass centre of the ground truth mask as the prompt.
    \item \textbf{SAM-randpt}: as known as SAM-random point, where we used two random points or coordinates inside the ground truth mask as the prompt.
\end{itemize}

In each of these approaches, SAM is directly utilised on the selected datasets, from infrared to X-ray security imagery, without any re-training or fine-tuning specific to those datasets. Moreover, all parameters are set to the default values \cite{kirillov2023segment}. When multiple masks corresponding to various regions or structures within the image are generated, we select the mask that exhibits the highest overlap with the ground-truth mask for evaluating the segmentation.



\section{Experimental Setup} \label{s:expset}
This section presents the used datasets and the implementation details of our experiments.
\vspace{-0.2cm}
\subsection{Datasets} \label{ss:db}
The following datasets are used in our evaluation: \\
\textbf{PIDray \cite{zhang2022pidray}}: this X-ray imagery dataset comprises a comprehensive collection of prohibited items, encompassing $12$ distinct classes: \textit{baton, bullet, gun, hammer, handcuffs, knife, lighter, pliers, power bank, scissors, sprayer, and wrench}. With a total of $29,457$ training images sourced from various environments, including airports, subway stations, and railway stations, this dataset offers a diverse and realistic representation of real-world scenarios. \\
\textbf{CLCXray \cite{zhao22:clcxray}}: this X-ray imagery dataset offers a substantial dataset featuring overlapping objects sourced from real-life scenarios, with a particular emphasis on hazardous liquids, thus broadening the scope of threat object research. The dataset comprises $7,652$ X-ray training images, a combination of real subway scenes and synthetically generated through manual bag design simulations. It encompasses $12$ categories, including five types of cutters (\textit{blade, dagger, knife, scissors, swiss army knife}) and seven types of liquid containers (\textit{can, carton drink, glass bottle, plastic bottle, vacuum cup, spray can, tin}). \\
\textbf{DBF6 \cite{akcay2018ganomaly}}: this dataset comprises conventional pseudo-colour X-ray security images captured by a Smith Detection dual-energy scanner, featuring four views. It includes six object classes: \textit{firearm, firearm part, knife, camera, ceramic knife, and laptop}, with a total of $8,100$ training images. Each object is meticulously annotated with segmentation masks across all views, enabling accurate identification, and is assigned a local index for seamless tracking. The dataset encompasses images depicting single objects as well as complex scenarios with multiple objects, providing diverse and challenging samples for analysis and training. \\
\textbf{FLIR \cite{flir2019flir}}: this infrared imagery dataset offers meticulously annotated single-channel grayscale infrared images covering various object classes. These images are captured in clear-sky conditions, encompassing both day ($60\%$) and night ($40\%$) settings. The Infrared imagery is acquired using a FLIR Tau2 camera, renowned for its Long Wave Infrared Cameras (LWIR), with a high resolution of $640 \times 512$ pixels. For our experiments, we primarily focus on the training set (totalling $7,859$ images) with three key object classes: \textit{Person, Bicycle, and Car}.

\begin{table*}[htb!]
\centering
\renewcommand*{\arraystretch}{1}
\fontsize{9pt}{9pt}\selectfont
\caption{PIDray: Average Recall comparison using IoU types: \{\textit{Bbox}\}  with various IoU thresholds.}
\vspace{-0.2cm}
\addtolength{\tabcolsep}{-3.50pt} 

\begin{tabular}{clcccccc}
\toprule
& Prompt & AR\tiny{[IoU=0.50:0.95]} & AR\tiny{[IoU=0.50]} & AR\tiny{[IoU=0.75]} & AR$_{S}$\tiny{[IoU=0.50:0.95]} &  AR$_{M}$\tiny{[IoU=0.50:0.95]} & AR$_{L}$\tiny{[IoU=0.50:0.95]}\\ \midrule  
\multirow{3}{*}{\rotatebox[origin=c]{90}{\scriptsize {\it {Bbox}}}} & {\it  $\rotatebox[origin=c]{180}{$\Lsh$}$bbox} & \textbf{0.767} & \textbf{0.972} & \textbf{0.855} & \textbf{0.767} & - & - \\
& {\it  $\rotatebox[origin=c]{180}{$\Lsh$}$centroid} & 0.393 & 0.613 & 0.401 & 0.393 & - & - \\
& {\it  $\rotatebox[origin=c]{180}{$\Lsh$}$randpt} &  0.456 & 0.687 & 0.469 & 0.456 & - & - \\  
\bottomrule

\end{tabular}%
\addtolength{\tabcolsep}{1pt}
\label{Table:mAP_pid}
\end{table*}

\begin{table*}[htb!]
\centering
\renewcommand*{\arraystretch}{1}
\fontsize{9pt}{9pt}\selectfont
\caption{CLCXray: Average Recall comparison using IoU type: \{\textit{Bbox}\} with various IoU thresholds.}
\vspace{-0.2cm}
\addtolength{\tabcolsep}{-3.50pt} 

\begin{tabular}{clcccccc}
\toprule
& Prompt & AR\tiny{[IoU=0.50:0.95]} & AR\tiny{[IoU=0.50]} & AR\tiny{[IoU=0.75]} & AR$_{S}$\tiny{[IoU=0.50:0.95]} &  AR$_{M}$\tiny{[IoU=0.50:0.95]} & AR$_{L}$\tiny{[IoU=0.50:0.95]}\\ \midrule  
\multirow{3}{*}{\rotatebox[origin=c]{90}{\scriptsize {\it {Bbox}}}} & {\it  $\rotatebox[origin=c]{180}{$\Lsh$}$bbox} & \textbf{0.797} & \textbf{0.992} & \textbf{0.894} & \textbf{0.704} & \textbf{0.697} & \textbf{0.801} \\
& {\it  $\rotatebox[origin=c]{180}{$\Lsh$}$centroid} & 0.597 & 0.821 & 0.644 & 0.512 & 0.477 & 0.596 \\
& {\it  $\rotatebox[origin=c]{180}{$\Lsh$}$randpt} &  0.282 & 0.423 & 0.292 & 0.544 & 0.295 & 250 \\  
\bottomrule

\end{tabular}%
\addtolength{\tabcolsep}{1pt}
\label{Table:mAP_clc}
\end{table*}




\begin{table*}[htb!]
\centering
\renewcommand*{\arraystretch}{1}
\fontsize{9pt}{9pt}\selectfont
\caption{DBF6: Average Recall comparison using IoU types: \{\textit{Bbox, Segm}\} with various IoU thresholds.}
\vspace{-0.2cm}
\addtolength{\tabcolsep}{-3.50pt} 

\begin{tabular}{clcccccc}
\toprule
& Prompt & AR\tiny{[IoU=0.50:0.95]} & AR\tiny{[IoU=0.50]} & AR\tiny{[IoU=0.75]} & AR$_{S}$\tiny{[IoU=0.50:0.95]} &  AR$_{M}$\tiny{[IoU=0.50:0.95]} & AR$_{L}$\tiny{[IoU=0.50:0.95]}\\ \midrule  
\multirow{3}{*}{\rotatebox[origin=c]{90}{\scriptsize {\it {Bbox}}}} & {\it  $\rotatebox[origin=c]{180}{$\Lsh$}$bbox} & \textbf{0.660} & \textbf{0.978} & \textbf{0.726} & \textbf{0.449} & \textbf{0.621} & \textbf{0.745}  \\
& {\it  $\rotatebox[origin=c]{180}{$\Lsh$}$centroid} & 0.399 & 0.703 & 0.398 & 0.281 & 0.364 & 0.460 \\
& {\it  $\rotatebox[origin=c]{180}{$\Lsh$}$randpt} & 0.400 & 0.696 & 0.400 & 0.249 & 0.314 & 0.475  \\  

\midrule

\multirow{3}{*}{\rotatebox[origin=c]{90}{\scriptsize {\it {Segm}}}} & {\it  $\rotatebox[origin=c]{180}{$\Lsh$}$bbox} & \textbf{0.537} & \textbf{0.912} & \textbf{0.533} & \textbf{0.320} & \textbf{0.474} & \textbf{0.572}  \\
& {\it  $\rotatebox[origin=c]{180}{$\Lsh$}$centroid} & 0.394 & 0.715 & 0.387 & 0.229 & 0.346 & 0.427 \\
& {\it  $\rotatebox[origin=c]{180}{$\Lsh$}$randpt} & 0.432 & 0.742 & 0.441 & 0.224 & 0.195 & 0.459   \\  

\bottomrule

\end{tabular}%
\addtolength{\tabcolsep}{1pt}
\label{Table:mAP_dbf6}
\end{table*}



\begin{table*}[htb!]
\centering
\renewcommand*{\arraystretch}{1}
\fontsize{9pt}{9pt}\selectfont
\caption{FLIR: Average Recall comparison using IoU type: \{\textit{Bbox}\} with various IoU thresholds.}
\vspace{-0.2cm}
\addtolength{\tabcolsep}{-3.50pt} 

\begin{tabular}{clcccccc}
\toprule
& Prompt & AR\tiny{[IoU=0.50:0.95]} & AR\tiny{[IoU=0.50]} & AR\tiny{[IoU=0.75]} & AR$_{S}$\tiny{[IoU=0.50:0.95]} &  AR$_{M}$\tiny{[IoU=0.50:0.95]} & AR$_{L}$\tiny{[IoU=0.50:0.95]}\\ \midrule  
\multirow{3}{*}{\rotatebox[origin=c]{90}{\scriptsize {\it {Bbox}}}} & {\it  $\rotatebox[origin=c]{180}{$\Lsh$}$bbox} & \textbf{0.606} & \textbf{0.991} & \textbf{0.627} & \textbf{0.546} & \textbf{0.688} & \textbf{0.784}  \\
& {\it  $\rotatebox[origin=c]{180}{$\Lsh$}$centroid} & 0.286 & 0.606  & 0.239 & 0.266 & 0.303 & 0.226 \\
& {\it  $\rotatebox[origin=c]{180}{$\Lsh$}$randpt} & 0.282 & 0.565 & 0.250 & 0.231 & 0.353 & 0.322  \\  

\bottomrule

\end{tabular}%
\addtolength{\tabcolsep}{1pt}
\label{Table:mAP_flir}
\end{table*}

\subsection{Implementation Details} \label{ss:implementation}
\vspace{-0.2cm}
We use the original implementation of SAM \cite{kirillov2023segment} with a ViT-H \cite{dosovitskiy2020image} backbone without further training or fine-tuning to assess its performance for non-visible band datasets. We evaluate on the training partition of each dataset since they allow for more statistically significant results. For each experiment, the prompts (\textit{bbox, centroid} and \textit{randpt}) are obtained from the ground truth datasets. The centroid is calculated as the mean of all ground truth mask vertices while the random points are obtained via Monte Carlo sampling of points within the bounding box and testing whether these points lie inside the polygon defining the ground truth mask until the desired number of random points is achieved (in our experiments, two random points). We report bounding box average recall (AR) in all our experiments by comparing the bounding box from the predicted mask against the ground truth bounding boxes. For DBF6, segmentation AR is also reported since ground truth masks are available. Additionally, we report the Recall for different intersection-over-union (IoU) thresholds and the mean IoU for each prompt/dataset pairs to evaluate the quality of the predictions. All experiments were run using an NVIDIA 3090Ti GPU. 
\section{Results} \label{ss:results}

\begin{figure*}[t]
\centering
    \includegraphics[width=5.5cm]{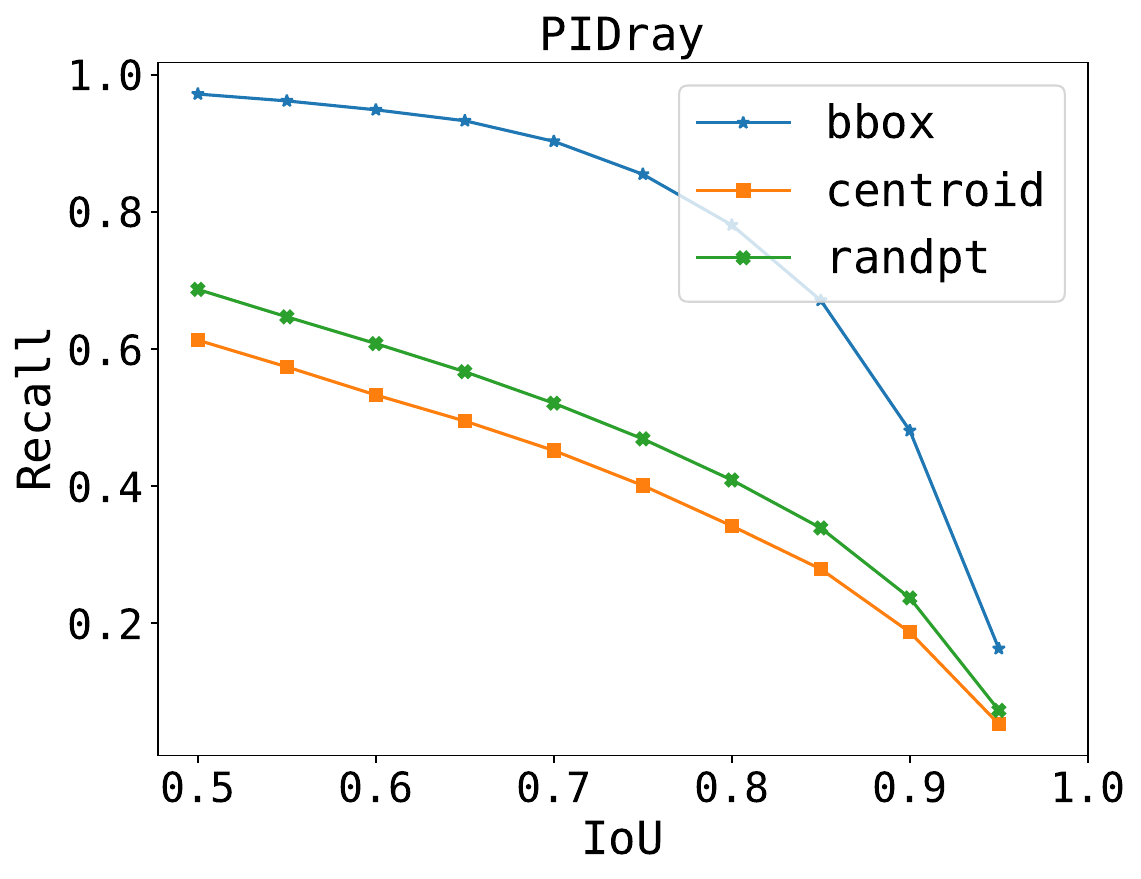}
    \hfill
    \includegraphics[width=5.5cm]{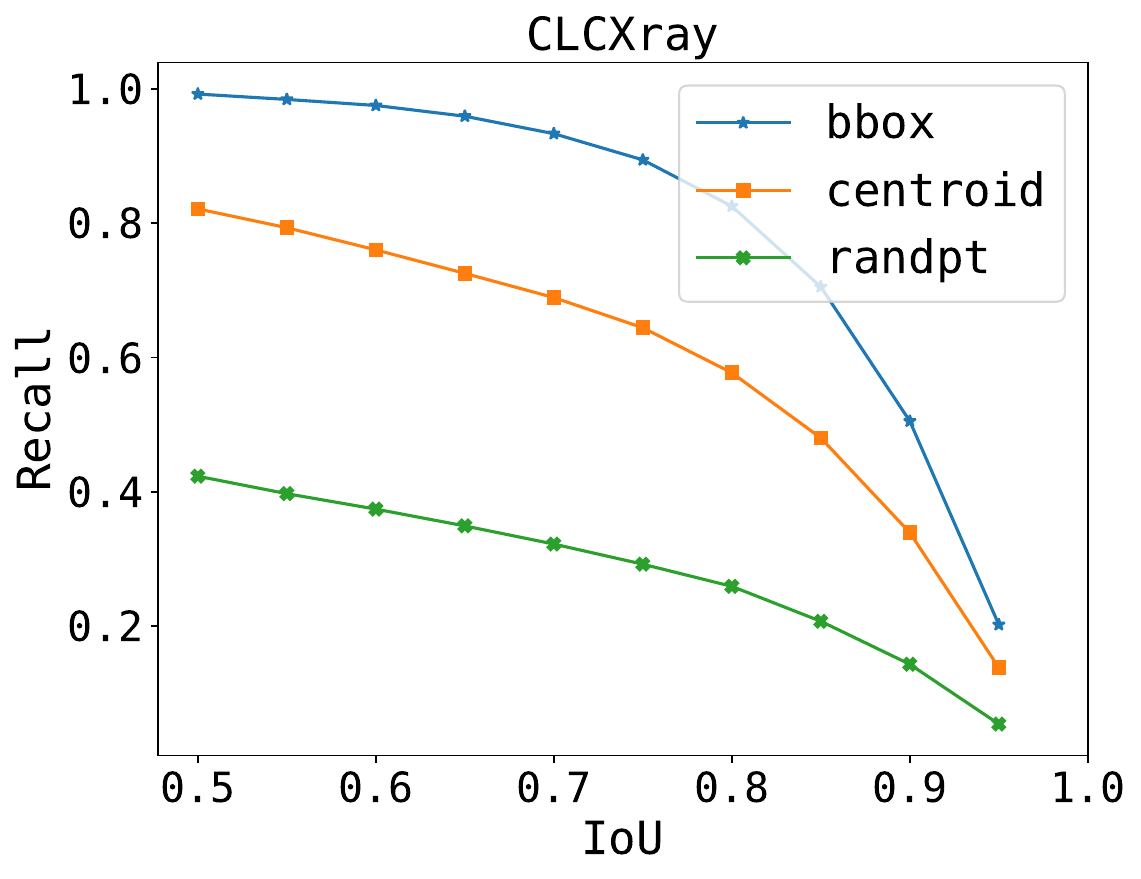}
    \hfill
    \vspace{-0.4cm}
    \includegraphics[width=5.5cm]{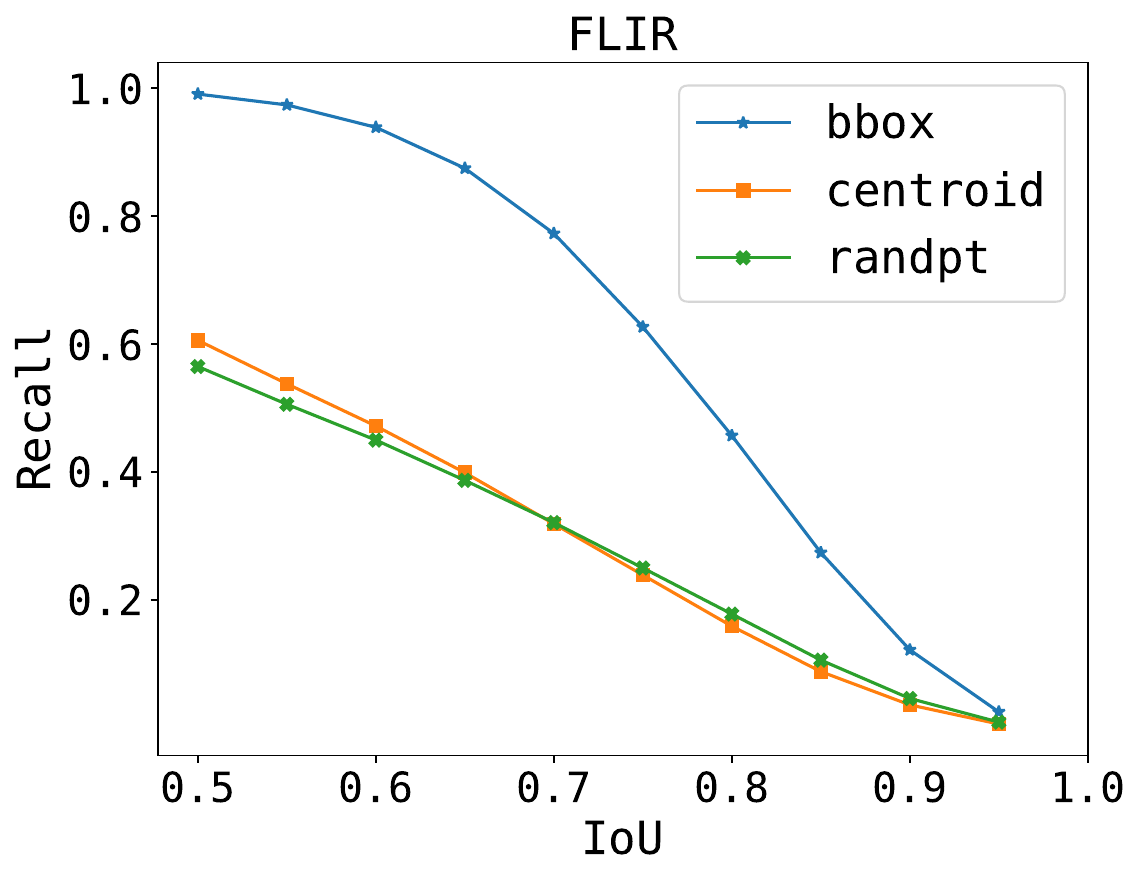}
    \caption{Recall performance using variational prompting strategies across different IoU thresholds and IoU type: \textit{Bbox}.}
    \label{fig:recall_iou}
\end{figure*}

 The resulting metrics for each dataset are summarised in Table \ref{Table:mAP_pid} to \ref{Table:mAP_flir}. We compiled our performance on the datasets across X-ray and infrared imagery, with varying IoU thresholds. The results are reported in terms of AR in two modes, bounding box (Bbox) mode and segmentation mask (Segm) mode. Note that only the DBF6 dataset has segmentation mask ground truth, meanwhile, the other three public datasets chosen only provide bounding box ground truth.  The results are reported under three prompts, (\textit{bbox}, \textit{centroid}, \textit{randpt}), as explained in Section \ref{sam_app}.

Having compiled a dataset across X-ray and infrared imagery, we noted that the segmentation efficacy of SAM is quantitatively influenced by the chosen prompting technique. For instance, across all datasets, bbox prompt shows superior results on object segmentation tasks on all IoU evaluation metrics, indicating that bbox prompting allows strong features combination within that particular area, by covering the entire object, making it more efficient in these instances.

We compare the result of the AR according to IoU evaluation criteria as shown in  Table \ref{Table:mAP_pid} and \cref{fig:recall_iou} (\textit{left}) for PIDray dataset \cite{zhang2022pidray}. For each of the given prompts, it consistently shows that AR decreases as IoU thresholds become stricter. At lower IoU thresholds, it is easier for SAM to have high AR because the criteria for correct detection are more lenient. As the IoU threshold increases, requiring more precise overlap, SAM ability to capture all class targets without also increasing false positives becomes more challenging, leading to lower AR.
It can be seen that whilst bbox prompt gives higher results, randpt gives slightly better results than the centroid point. The distinct class of PIDray, as explained in Section \ref{ss:db}, give more advantages to SAM with randpt prompt, where having more points inside the target class is effectively better at segmenting compared to centroid point, with a small margin.

We further analyse the AR based on IoU evaluation criteria, as presented in Table \ref{Table:mAP_clc} and \cref{fig:recall_iou} (\textit{middle}) for the CLCXray dataset \cite{zhao22:clcxray}. This analysis reveals a consistent trend with the PIDray results; however, the centroid prompt notably outperforms the random point prompt by a significant margin, unlike the PIDray findings (\cref{fig:recall_iou} (\textit{left})). 
In this regard, the critical influence of material composition may influence the effectiveness of prompt selection. The CLCXray dataset has a class-wise combination which consists of organic material (carton drinks, plastic bottles) as well as metallic material (spray cans, tin), contrasting with the PIDray dataset that exclusively consists of metallic composition, as discussed in Section \ref{ss:db}. The presence of clutter, especially involving organic materials, poses greater challenges for the random point prompt due to the less distinctive features of these objects. This suggests that increasing the number of prompt points does not automatically enhance segmentation performance. Instead, strategically placing a prompt at the centre of the target object significantly improves results, as evidenced in \cref{fig:recall_iou} (\textit{middle}).

\begin{figure}[tb!]
\centering
    \includegraphics[width=4.2cm]{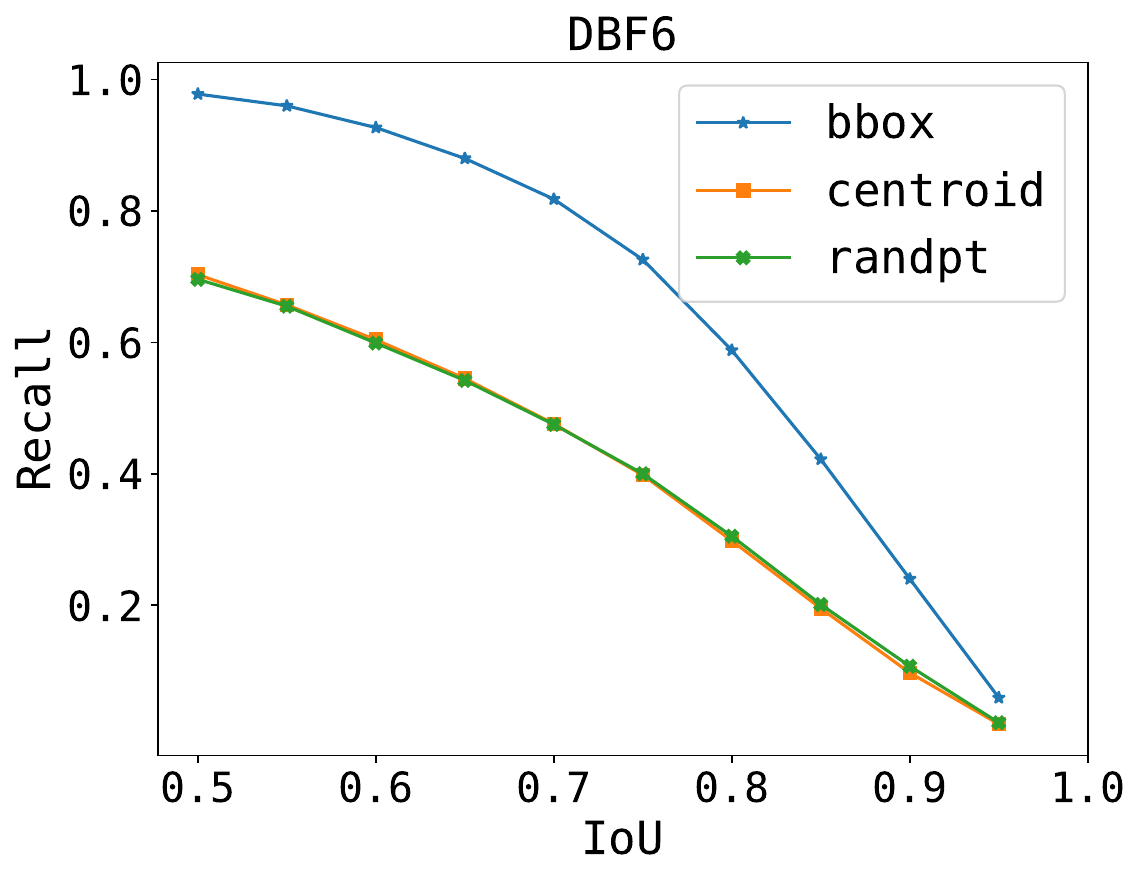}
    \hspace{-0.24cm}
    \includegraphics[width=4.2cm]{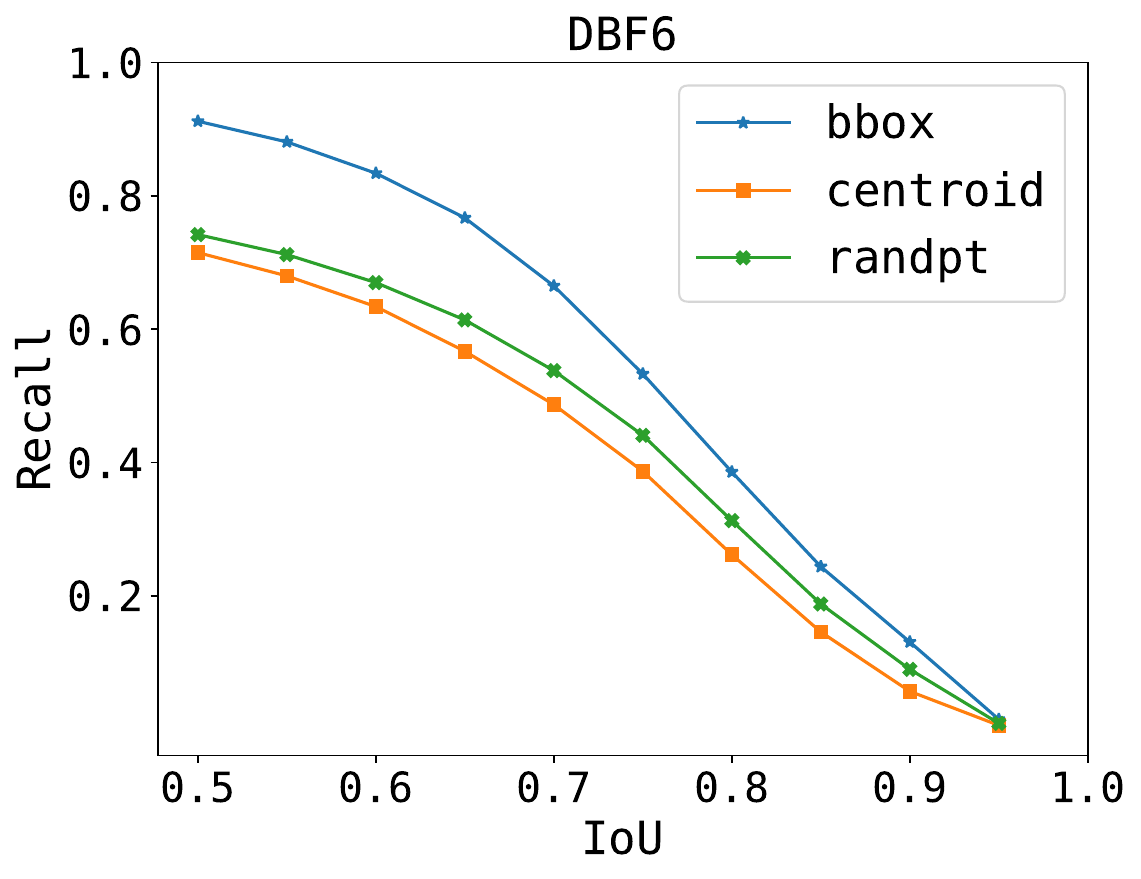}
    \vspace{-0.8cm}
    \caption{DBF6: Recall performance using variational prompting strategies across different IoU thresholds and IoU types: \textit{Bbox} (left), \textit{Segm} (right).}
    \label{fig:recall_iou_dbf6}
\end{figure}

In our final analysis of X-ray imagery, we focus on the DBF6 dataset results, detailed in Table \ref{Table:mAP_dbf6} and illustrated in \cref{fig:recall_iou_dbf6}. This evaluation considers two modes: bounding box (Bbox) and segmentation (Segm). While the bbox prompt maintains consistent performance in the Bbox mode, aligning with the patterns observed in \cref{fig:recall_iou}, we notice a marginal decline in its effectiveness in the Segm mode. Although SAM can generate bounding boxes from segmentation masks, this approach proved to be less efficient, primarily due to difficulty associating the bbox prompt with Segm mode of SAM. Conversely, the point-based prompting methods, both centroid and randpt, demonstrated stable performance across both modes, differing only slightly. This consistency indicates the robust adaptability of point prompts to varying segmentation tasks within X-ray imagery analysis.

In our analysis of infrared imagery, as presented in Table \ref{Table:mAP_flir} and depicted in \cref{fig:recall_iou} (\textit{right}), we observe that while the bbox prompt generally results in higher AR, there is a notable decrease in AR performance as the IoU threshold increases for all types of prompts. This trend suggests SAM's limited ability to understand the characteristics of class-specific infrared imagery, given its training predominantly on natural images. We propose that fine-tuning the SAM model with an infrared imagery dataset could significantly enhance its segmentation accuracy and effectiveness, providing more robust quantitative results.

\begin{table}[tb!]
\centering
\renewcommand*{\arraystretch}{1}
\fontsize{9pt}{9pt}\selectfont
\caption{Mean IoU for each prompt/dataset pairs.}
\vspace{-0.2cm}
\addtolength{\tabcolsep}{-3.50pt} 
\resizebox{\linewidth}{!}{%
\begin{tabular}{lccccc}
\toprule
Prompt & DBF6 (Box) & DBF6 (Segm) & PIDRay & CLCXray &  FLIR \\ \midrule  
{\it  $\rotatebox[origin=c]{180}{$\Lsh$}$bbox} & $0.808 \pm 0.130$ & $0.730 \pm 0.171$ & $0.849 \pm 0.131$ & $0.870 \pm 0.104$ & $0.779 \pm 0.110$ \\
{\it  $\rotatebox[origin=c]{180}{$\Lsh$}$centroid} & $0.603 \pm 0.288$ & $0.591 \pm 0.287$ & $0.577 \pm 0.299$ & $0.728 \pm 0.255$ & $0.534 \pm 0.235$ \\
{\it  $\rotatebox[origin=c]{180}{$\Lsh$}$randpt} & $0.606 \pm 0.283$ & $0.621 \pm 0.281$ & $0.634 \pm 0.280$ & $0.440 \pm 0.337$ & $0.547 \pm 0.262$ \\  
\bottomrule

\end{tabular}%
 }
\addtolength{\tabcolsep}{1pt}
\label{Table:miou}
\vspace{-0.4cm}
\end{table}
\begin{figure}[tb!]
\centering
    \includegraphics[width=0.9\linewidth]{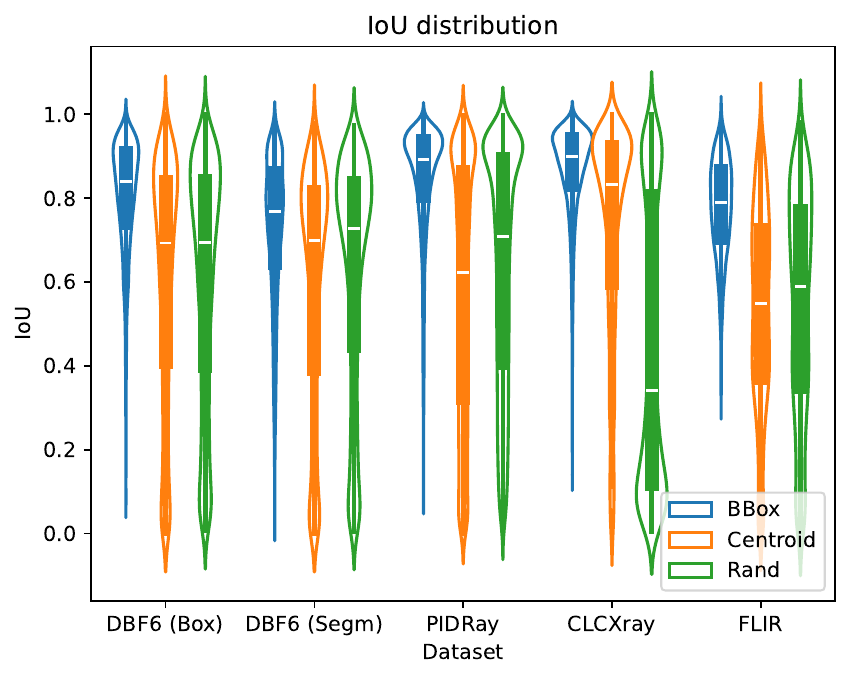}
    \vspace{-0.5cm}
    \caption{IoU distribution for each prompt/dataset pair.}
    \label{fig:iou_distribution}
    \vspace{-0.4cm}
\end{figure}

\begin{figure}[tb!]
\centering
    \includegraphics[width=\linewidth]{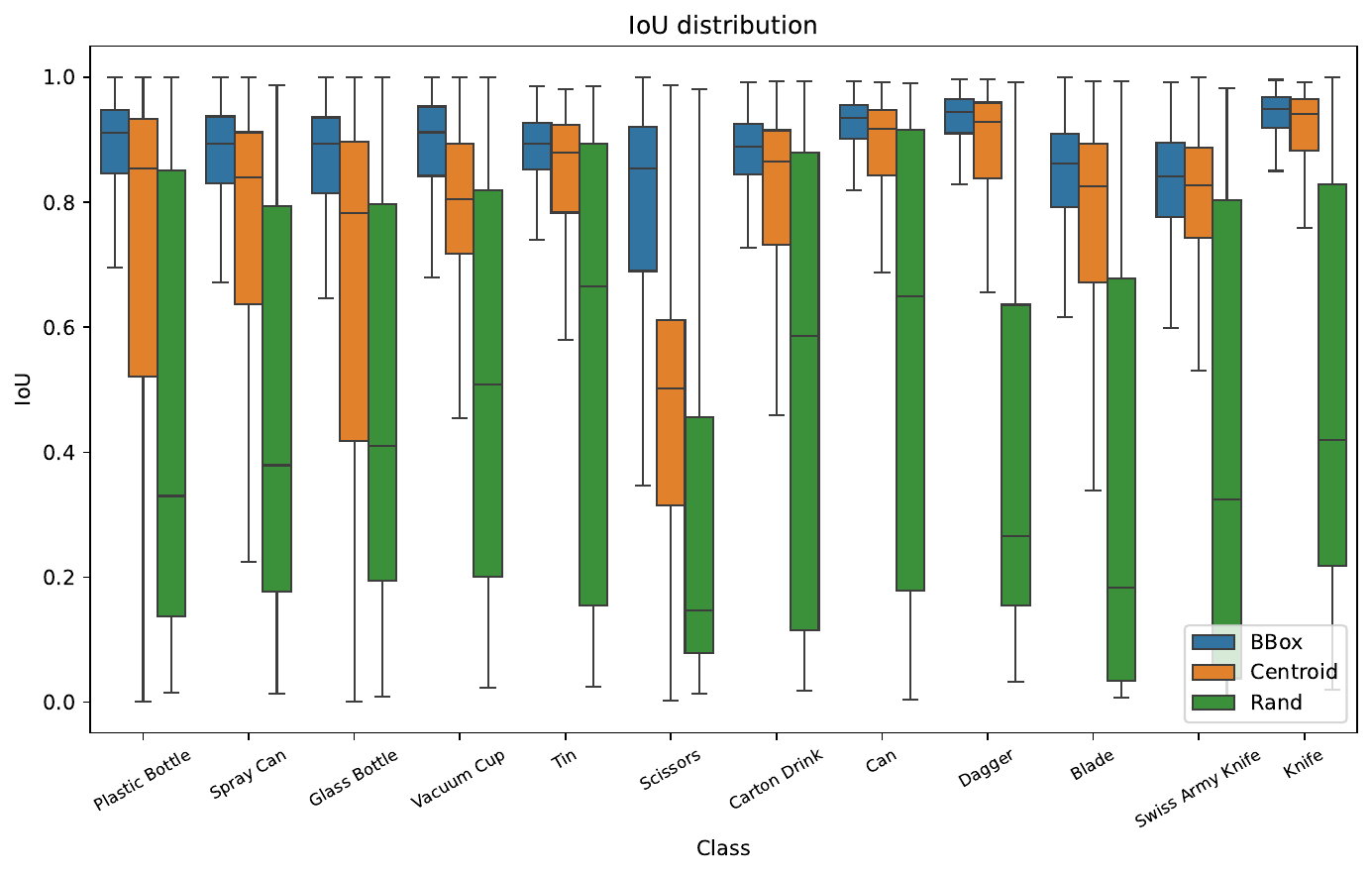}
    \vspace{-0.8cm}
    \caption{Class-wise IoU distribution on the CLCXray dataset.}
    \vspace{-0.4cm}
    \label{fig:clc_iou_distribution}
\end{figure}

\begin{figure*}[tb!]
\centering
    \includegraphics[width=\textwidth]{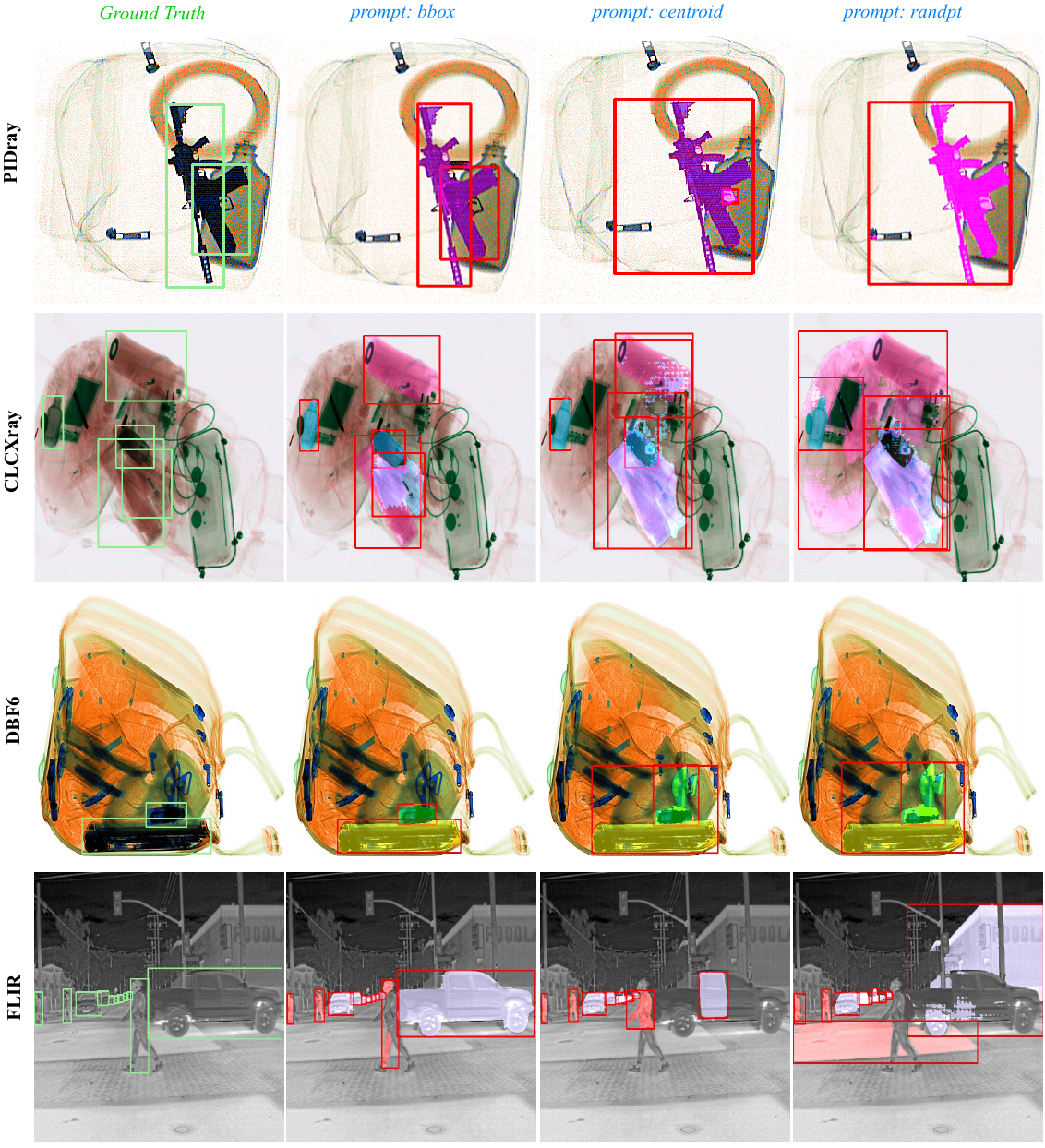}
    
    \caption{The segmentation results obtained by SAM, utilising variational prompting strategies, are examined across PIDray, CLCXray, DBF6, and FLIR datasets. Notably, the prompt bbox consistently yields the most accurate segmentations. However, the other two prompting strategies occasionally encounter challenges, particularly in scenarios where objects are overlapped and cluttered, as observed in the X-ray datasets.}
    \label{fig:qe1}
\end{figure*}

Table \ref{Table:miou} presents the mean IoU for each prompt/dataset pair. Overall, it is observed that the bounding box prompts usually lead to a good bounding box prediction, meaning that SAM can segment the object inside the proposed bounding box. This is still confirmed for the DBF6 segmentation dataset, with a relatively high mean mask IoU. Among the bounding box datasets, FLIR obtains the lowest mean IoU, indicating its challenging nature to SAM, which is explained by the low contrast of the objects against the background (see \cref{fig:qe1}). From the centroid and random point prompts, it is observed that, generally, two random points lead to a better performance than the centroid. While this might seem counter-intuitive, it indicates that the centroid is not always the most significant point, while two random points within the object lead to more cues for SAM. The contrary is noted for the CLCXray, where several objects consist of thin objects and bottles containing liquids \cite{zhao22:clcxray}, which are difficult to capture by an X-ray machine and two random points might still be confused with the background. These trends are further confirmed in the IoU distribution shown in \cref{fig:iou_distribution}. It is noted that the three types of prompts generally yield a good segmentation mask for DBF6 and PIDRay, where the test objects are metallic. On the other hand, the CLCXray shows a lower performance with a high density of low IoU objects for the random points prompt. To further investigate this, \cref{fig:clc_iou_distribution} shows the class-wise IoU distribution of the CLCXray dataset. It is seen that the best-performing classes for the random point prompt are the tin and the cans, which are metallic and easily segmented. On the other hand, the worst performances are obtained for thin objects (scissors, blades and daggers) and organic material objects (such as plastic bottles), where the choice of the random point might significantly impact the prediction. Finally, it is also observed that SAM has the poorest performance on the FLIR dataset when using point prompts, which is again attributed to the low contrast of the dataset.

The qualitative analysis, which examines the variance in prompts and their alignment with ground truths across different datasets, is illustrated in \cref{fig:teaser,fig:qe1}. For the PIDray dataset (\cref{fig:qe1}, \textit{$1^{st}$ row}), we observe that point-based prompts often extend beyond the actual object boundaries, leading to an increased occurrence of false positives compared to the bounding box prompts. In the case of the CLCXray dataset (\cref{fig:qe1}, \textit{$2^{nd}$ row}), which features a high degree of overlap between organic and metallic materials, strategically placing a prompt directly at the centre of the target significantly improves bounding box precision compared to random placement. The randpt prompt particularly struggles with smaller objects, often exceeding their boundaries. Regarding the DBF6 dataset (\cref{fig:qe1}, \textit{$3^{rd}$ row}), point-based prompts generally achieve better segmentation than bounding box prompts, although they tend to generate more false positives for classes with smaller objects. In the FLIR dataset (\cref{fig:qe1}, \textit{$4^{th}$ row}), while the bounding box prompts offer superior outcomes, point-based prompts fail to accurately localise distinct classes such as pedestrians and cars, likely due to the significant domain shift between the infrared imagery encountered here and the visible band imagery upon which SAM was trained.
\vspace{-0.2cm}\section{Conclusion}
This work presents a thorough assessment of the Segment Anything Model (SAM) performance for images beyond the visible spectrum. We evaluate SAM within three types of prompts, namely bounding box, centroid and random points, on three X-ray security imagery datasets and an infrared surveillance dataset. Our results suggest that while SAM exhibits a great capability to segment objects when given a bounding box prompt, its performance drops when given point prompts. Specifically, it is observed that SAM extends objects beyond their boundaries in X-ray images, with particular difficulty for objects based on organic materials (which might be confused with the background). Additionally, the low-contrast characteristic of infrared images and the different appearance of the objects impose a significant challenge on SAM, with poor segmentation performance using the centroid and random point prompts. Future directions may include fine-tuning SAM to the assessed image modalities to increase its segmentation performance. This would allow to streamlining the dataset annotation processes, reducing the reliance on manual labelling. Such advancements could facilitate the creation and enrichment of datasets across various image modalities, significantly broadening the scope and utility of machine learning applications in areas where data collection and annotation have traditionally posed challenges.
\vspace{01cm}

\hfill

\begingroup
\setstretch{0.92}
{\small
\bibliographystyle{ieee_fullname}
\bibliography{egbib,object_detection,ref_mod,ref_1,IEEEabrv,IEEEreference,advattak,eccv_ref}
}

\endgroup

\end{document}